\documentclass{article}

\usepackage[preprint, nonatbib]{neurips_2026}

\usepackage[utf8]{inputenc} %
\usepackage[T1]{fontenc}    %
\usepackage[hidelinks]{hyperref}       %
\usepackage{url}            %
\usepackage{booktabs}       %
\usepackage{amsfonts}       %
\usepackage{nicefrac}       %
\usepackage{microtype}      %
\usepackage{xcolor}         %
\usepackage{graphicx}       %
\usepackage{wrapfig}

\title{LUCoS: Latent Unsupervised Context Selection for Tabular Foundation Models}

\author{
Oroel Ipas$^{1}$\\
\texttt{oroel@ugr.es} \\
\And
Guillermo Gomez-Trenado$^{1}$\\
\texttt{guillermogomez@ugr.es}\\
\And
Rocío Romero-Zaliz$^{2,1,3,4}$\\
\texttt{rocio@ugr.es}\\
\And
Isaac Triguero$^{2,1}$\\
\texttt{triguero@decsai.ugr.es}\\[1.0em]
$^1$Andalusian Research Institute in Data Science and Computational Intelligence (DaSCI)\\
$^2$Department of Computer Science and Artificial Intelligence (DECSAI),\\
$^3$Research Center in Information and Communication Technologies (CITIC),\\
$^4$Instituto de Investigación Biosanitaria Ibs.GRANADA, \\
University of Granada, Granada, 18071, Spain
}

\newcommand{\codeurl}{\url{https://github.com/ari-dasci/S-LUCoS}}
\usepackage{subcaption}

\usepackage{makecell}

\usepackage{amsmath}
\usepackage{multirow}
\usepackage{cleveref}

\begin{document}

\maketitle

\vspace{-9pt}

\begin{abstract}

Selecting which instances to label is a key challenge in low-label tabular learning. For recent Tabular Foundation Models such as TabPFN, context selection directly determines predictive performance. Supervised oracle experiments show that carefully chosen labeled context sets can strongly outperform random selection under the same labeling budget. However, the cold-start setting, where instances must be selected before any labels are available, has received little attention in the TFM literature. This problem is fundamentally geometric. In vision and language, foundation models induce embedding spaces where simple geometric selection methods are effective. In contrast, tabular instance selection has so far been performed predominantly in the original tabular space, which lacks a natural metric; heterogeneous types, mixed scales, and nonlinear interactions make raw-space distances unreliable for context construction, and original-space selection falls below random on the majority of datasets as the budget grows. We propose \textbf{LUCoS} (\textbf{L}atent \textbf{U}nsupervised \textbf{Co}ntext \textbf{S}election), which replaces raw-feature geometry with the latent geometry induced by embeddings from an unsupervised Prior-Fitted Network (PFN) and selects representative medoids as context. Evaluated on 67 OpenML-CC18 datasets across six low-label budgets, LUCoS ranks first under mean AUC, ACC, and F1, with conclusions stable across metrics and dataset-level robustness checks. A gain decomposition reveals a simple mechanism: at the smallest budgets, the main benefit comes from enforcing coverage; as the budget increases, the decisive factor becomes the representation space in which coverage is measured. LUCoS mitigates failures of original feature space selection, showing that reliable unsupervised context selection depends less on selector sophistication than on defining representativeness in a meaningful representation geometry.

\end{abstract}

\section{Introduction}

Tabular data—structured in rows and columns with heterogeneous feature types—is the dominant data format across domains such as healthcare, finance, energy, manufacturing, and industrial systems~\cite{breugel2024position, grinsztajn2025tabpfn}. Yet, in practice, labeled tabular data is often scarce: annotations may require costly expert involvement, regulated access, or time-consuming collection pipelines~\cite{borisov2022deep}. This makes low-label regimes common in applied settings and raises a practically important question: \textbf{given a fixed labeling budget, which training instances should be annotated?}

\begin{figure}[t]
  \centering
  \includegraphics[width=1\linewidth]{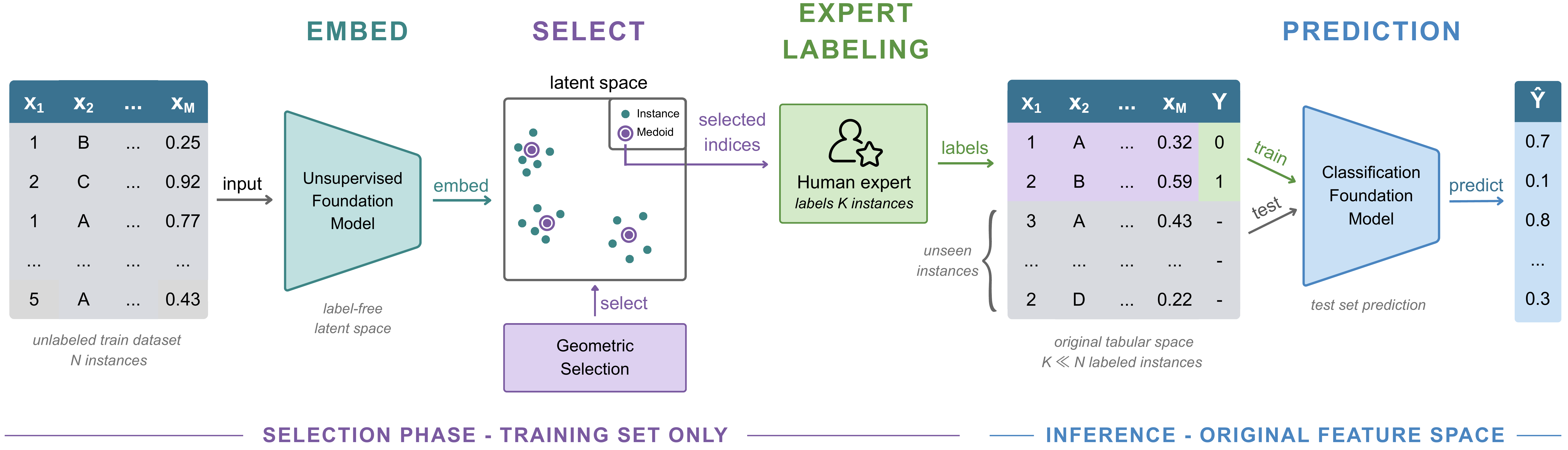}
  \caption{\textbf{Overview of LUCoS.} The method consists of four stages: \textbf{(i)} embedding the unlabeled training data into a high-dimensional latent representation space, \textbf{(ii)} selecting representative instances using a geometric criterion, \textbf{(iii)} mapping the selected instances back to the original tabular space and querying the oracle to label them, and \textbf{(iv)} using the labeled instances as the in-context set of a supervised TFM for prediction on unseen data. Instance selection is performed without label access; test instances are never used during embedding, selection, or subset construction; they are reserved exclusively for final evaluation.}
  \label{fig:graphical_abstract}
\end{figure}

Recent Tabular Foundation Models (TFMs) have emerged as state-of-the-art approaches for low-data tabular supervised learning, with representative models including TabPFN~\cite{hollmann2023tabpfn, hollmann2025accurate}, TabICL~\cite{qu2025tabicl, qu2026tabiclv2}, and TabDPT~\cite{ma2025tabdpt}. TFMs perform inference via In-Context Learning (ICL), conditioning predictions on a labeled context set rather than fitting task-specific parameters through gradient-based training. This suggests that they may naturally operate as effective few-shot learners, analogous to behavior observed in Large Language Models~\cite{brown2020language}, which has also been explored for few-shot tabular learning in approaches such as TabLLM~\cite{hegselmann2023tabllm}.

We empirically validate this hypothesis by evaluating TFMs in low- and extremely low-label regimes. Our results confirm not only strong few-shot performance but also high sensitivity to the specific instances selected as context, revealing substantial headroom for improvement via instance selection. This raises a natural question: can this headroom be partially addressed without access to labels?

Selecting instances without prior labels, known as the cold-start problem~\cite{jin2022cold}, has been extensively studied in computer vision, including instance selection~\cite{wang2022unsupervised, shao2024enhancing}, class discovery~\cite{mannix2023cold}, and coreset construction~\cite{sener2018active}. These approaches typically rely on distance-based methods that assume a meaningful geometry in the representation space, often induced by foundation models that align representations with semantic similarity (e.g., DINO~\cite{caron2021emerging}, MAE~\cite{he2022masked}, CLIP~\cite{radford2021learning}). However, label-free instance selection has received little attention in the context of TFMs. Unlike vision and language, tabular data lacks a generally reliable representation space~\cite{jiang2026representation, wang2025survey}: instance selection is typically performed in the original tabular space, where heterogeneous types, mixed scales, sparsity, missing values, and nonlinear interactions make distances poorly aligned with predictive similarity~\cite{koshil2024towards}. As a result, selection in the original feature space often leads to inconsistent behavior. Random sampling provides a simple baseline, but offers no control over coverage and may underrepresent sparse yet important regions of the input distribution.

In this work, we address this gap by replacing the geometry of the original feature space with embeddings induced by the encoder of an unsupervised TFM~\cite{zhao2026tabclustpfn}, enabling the construction of task-relevant representations for tabular data. We show that performing instance selection in this latent space allows simple geometric methods to identify informative context sets, which are then annotated and used as the TFM context for downstream prediction. We refer to this setting as \emph{unsupervised context selection}. In contrast to context optimization methods, which assume fully labeled datasets and focus on scalable inference~\cite{feuer2024tunetables,zabergja2026end}, our approach is entirely label-free and designed to improve predictive performance under low-label budgets.

The resulting approach is lightweight, label-free, and integrates naturally into the TFM inference pipeline. Since ICL inference scales quadratically with context size, selecting compact informative subsets also reduces the $\mathcal{O}(N^2)$ cost of ICL inference as a byproduct. Our contributions are:

\begin{enumerate}

\item We show that TFM performance in low-label regimes is highly sensitive to the selected context instances, motivating instance selection as a key component of tabular ICL.
\item We propose LUCoS, a principled unsupervised context selection pipeline that operates in the latent space of prior-fitted clustering models, enabling simple geometric methods such as K-Medoids to identify high-performing subsets for downstream classification.
\item We show that LUCoS ranks first across all six low-label budgets on 67 OpenML-CC18 datasets, and decompose its gain into selection and representation components. The analysis reveals a regime shift in which the latent representation becomes the dominant driver of performance beyond extremely low-label budgets, rescuing datasets where original-space selection collapses below random.
    
\end{enumerate}

We provide the implementation of the proposed method, along with the experimental scripts required to reproduce our results, at \codeurl.

\section{Related work}

\paragraph{Few-shot and semi-supervised learning.}
Few-shot learning aims to generalize from a small number of labeled examples, while semi-supervised learning additionally leverages unlabeled data during training. Semi-supervised learning has been extensively studied in computer vision, tabular learning, and related domains~\cite{yang2022survey, van2020survey}. Recent work such as LoopTabFM~\cite{li2025theoretical} explores these ideas for TFMs. In contrast, while few-shot learning has been extensively studied in computer vision~\cite{wang2020generalizing}, it remains relatively underexplored for TFMs, particularly in the cold-start setting where instances must be selected for annotation before any labels are available.

\paragraph{Unsupervised instance selection.}

Unsupervised instance selection aims to identify informative subsets from unlabeled data under limited labeling budgets. Existing approaches such as USL~\cite{wang2022unsupervised}, RDSS~\cite{shao2024enhancing}, ZCore~\cite{griffin2026zero}, and core-set selection~\cite{sener2018active} rely on geometric criteria such as neighborhood structure, density, or distribution coverage. These methods assume that distances in the representation space are meaningful, so geometric diversity correlates with predictive utility. While this assumption often holds in computer vision when selection is performed on deep latent representations, we hypothesize that it is less reliable in the original tabular feature space. Existing tabular methods operate directly on raw features~\cite{garcia2012prototype}, where heterogeneous feature types, incompatible scales, missing values, sparsity, and nonlinear interactions can make distances poorly aligned with predictive similarity. Consequently, coverage or diversity criteria may select geometrically spread samples that are not informative for downstream prediction. This suggests that representation quality is a key bottleneck for tabular instance selection. %

\paragraph{Prior-fitted networks (PFNs).}

PFNs are a class of TFMs pretrained to approximate Bayesian inference over distributions of synthetic and/or real tasks, enabling strong performance from very small labeled datasets. While foundation models in computer vision are predominantly pretrained through self-supervised learning, tabular foundation models have largely followed the PFN paradigm. Supervised PFNs such as TabPFN~\cite{hollmann2023tabpfn, hollmann2025accurate}, TabICL~\cite{qu2025tabicl, qu2026tabiclv2}, and TabDPT~\cite{ma2025tabdpt} perform prediction through ICL, conditioning directly on labeled context instances without gradient-based optimization. As a result, predictive performance becomes highly sensitive to the selected context set, making context construction a central challenge in low-label tabular learning.

\paragraph{Label-free representation learning for tabular data.}
Learning useful representations from unlabeled data has become a central objective in modern machine learning, enabling feature spaces where simple geometric operations are effective across domains such as computer vision (e.g., DINO~\cite{caron2021emerging}, MAE~\cite{he2022masked}) and natural language processing (e.g., BERT~\cite{devlin2019bert}). In tabular data, however, learning such representations remains challenging due to the absence of a natural spatial or sequential structure~\cite{wang2025survey}. Early approaches relied on pretext tasks and self-supervised objectives such as contrastive learning with data augmentations, including VIME~\cite{yoon2020vime}, SCARF~\cite{bahri2022scarf}, SAINT~\cite{somepalli2022saint}, and TransTab~\cite{wang2022transtab}. While effective, their applicability is often limited to pretraining domains~\cite{jiang2026representation}. More recent work has explored PFN-based approaches with synthetic data. In particular, ZEUS~\cite{marszaek2025zeus} introduces zero-shot embeddings for unsupervised clustering, while TabClustPFN~\cite{zhao2026tabclustpfn} performs Bayesian inference over cluster assignments and cluster counts in a single forward pass. These methods suggest that PFNs can induce meaningful label-free latent geometries for tabular data. However, their application to downstream tasks such as unsupervised instance selection remains largely unexplored.

\paragraph{Context optimization for TFMs.}
Because TFM predictions and computational cost both depend strongly on the context set, recent work has explored context optimization methods to improve efficiency and scalability. These approaches compress datasets into smaller informative contexts for inference. TuneTables~\cite{feuer2024tunetables} performs soft prompt tuning on labeled validation data, while TACO~\cite{zabergja2026end} generates compact synthetic contexts for efficient inference. However, these methods assume fully labeled datasets and mainly optimize context representations rather than selecting informative instances from unlabeled pools. Consequently, they are not applicable in cold-start settings.

In contrast to prior work, which either relies on labeled data for context optimization or performs selection in the original feature space, we propose to induce a meaningful latent geometry via prior-fitted models, enabling effective unsupervised instance selection under low-label budgets.

\section{Methodology} \label{sec:methodology}
TFMs consume a labeled context set at inference time, which makes few-shot performance sensitive to which instances populate that set: two labeled support subsets of the same size can yield substantially different predictions. We study \emph{unsupervised context selection}, where a small subset of training instances must be chosen for annotation before any labels are available, with the goal of maximizing downstream classification performance once the labeled subset is used as the in-context support set. As motivated in Section~\ref{subsec:exp_sensitivity}, this sensitivity creates substantial headroom for label-free selection.

\paragraph{Desiderata.} Unsupervised context selection imposes a set of constraints that distinguish it from active learning, label-aware coreset selection, and dataset distillation. A method must be \textbf{(i)} \emph{label-free}: neither the selection procedure nor the representations used during selection may depend on downstream task labels; \textbf{(ii)} \emph{test-blind}: test instances must remain unavailable during selection; \textbf{(iii)} \emph{instance-faithful}: the selected samples must correspond to real training instances, since synthetic prototypes cannot be sent to an oracle for annotation; and \textbf{(iv)} \emph{single-round}: selection must succeed without iterative annotation or feedback. Together, these constraints motivate a representation-driven geometric formulation.

\paragraph{Framework: representation $\times$ selector.} We treat unsupervised context selection as the composition of two factors: a \emph{representation map} $\phi : \mathcal{X} \to \mathcal{Z}$ that embeds raw instances into a geometry where similarity is meaningful, and a \emph{geometric selector} $\sigma$ that returns the indices of $K$ representative points in $\mathcal{Z}$. This decomposition is not merely presentational: holding one factor fixed while varying the other isolates the contribution of each component, which we exploit experimentally in Section \ref{subsec:exp_comparison}. \emph{Our central hypothesis is that the raw tabular feature space provides an unreliable geometry for unsupervised context selection, and that representations induced by unsupervised PFNs offer a more suitable one.} This makes unsupervised PFNs~\cite{zhao2026tabclustpfn, marszaek2025zeus} natural candidates for the representation map.
 
\subsection{LUCoS}
 
We propose \textbf{LUCoS} (\textbf{L}atent \textbf{U}nsupervised \textbf{Co}ntext \textbf{S}election), a four-stage pipeline that follows this framework, illustrated in Fig.~\ref{fig:graphical_abstract}. First, unlabeled training instances are projected into a latent representation space induced by an unsupervised PFN. Second, representative samples are selected in this latent space using a geometric coverage criterion. Third, only the selected samples are queried once from an oracle for annotation. Finally, the labeled subset, expressed in the original feature space, is used as the in-context set for prediction by a separate supervised TFM. The latent representation is used \emph{only} for selection; downstream inference uses the standard preprocessing pipeline of the supervised TFM. This separation isolates the effect of the selection strategy while keeping the downstream predictor fixed.
 
\paragraph{Component choices.} For $\phi$, we use the PIN Encoder module of TabClustPFN~\cite{zhao2026tabclustpfn}. We use only the encoder because the full model produces synthetic clustering prototypes and is constrained to at most 10 inferred clusters. The PIN Encoder instead provides a 512-dimensional embedding for each instance, independent of the number of clusters. For $\sigma$, we use $K$-Medoids under Euclidean distance, motivated by three properties that match the desiderata: (a) medoids correspond to real training instances rather than synthetic centroids or prototypes, so the selected set is annotatable (desideratum iii); (b) $K$-Medoids minimizes an explicit coverage objective, matching the few-shot requirement that a small support set must cover the data distribution; and (c) it is deterministic up to initialization, requires no class prior, and admits a number of medoids fixed exactly to the labeling budget $K$. The framework is not restricted to this particular pair and is compatible with any unsupervised representation model that preserves clusterable latent structure (Appendix \ref{app:vs_zeus}).
 
\paragraph{Formal definition.} Given a training dataset $\mathcal{D} = \{x_i\}_{i=1}^{N}$, we first compute latent embeddings
\begin{equation}
    z_i = \phi(x_i), \qquad z_i \in \mathbb{R}^{512}.
\end{equation}
We then perform $K$-Medoids clustering over $\{z_i\}_{i=1}^{N}$ using Euclidean distance, with the number of medoids fixed to the labeling budget $K$. The selected index set is given by
\begin{equation}
    \mathcal{I}_K =
    \operatorname*{argmin}_{\mathcal{I}\subseteq\{1,\dots,N\},\,|\mathcal{I}|=K}
    \sum_{i=1}^N \min_{j\in\mathcal{I}} \|z_i - z_j\|_2,
\end{equation}
yielding the context subset $\mathcal{Q}_K = \{x_i \mid i \in \mathcal{I}_K\}$. Because medoids correspond to real training instances, every element of $\mathcal{Q}_K$ traces back to the original feature space and can be sent to the oracle for annotation. Once labeled, $\mathcal{Q}_K$ is consumed as the in-context support set by the downstream supervised TFM, for which we use TabPFN-2.5~\cite{grinsztajn2025tabpfn}. Test instances are never accessed during embedding, selection, or subset construction.
 
\paragraph{From method to mechanism.} The framework yields two testable predictions. Holding the selector fixed, replacing the original feature space with a clusterable latent space should help in regimes where the latent geometry is informative. Holding the representation fixed, selectors that do not exploit the latent geometry should not benefit from being applied in it. In Section \ref{sec:experiments} we test both predictions by evaluating multiple combinations of representations and selectors, and we further decompose the LUCoS gain over random selection into a selector component and a representation component to identify which factor drives the improvement at each labeling budget.

\subsection{Benchmark and evaluation protocol}

We use TabPFN-2.5~\cite{grinsztajn2025tabpfn} as the downstream classifier due to its strong performance in low-data regimes and its in-context learning formulation. We evaluate on the OpenML-CC18 benchmark~\cite{bischl2021openml}, restricting the suite to datasets with at most ten classes due to the constraints of TabPFN-2.5. This filtering yields 67 out of 72 classification datasets covering numerical and categorical features, missing values, and a wide range of dataset sizes.
For all methods, numerical missing values are imputed using the mean of the training split, while categorical missing values are imputed using the mode, following standard practice~\cite{bahri2022scarf}.

Let \(C\) denote the number of classes. We evaluate exponentially increasing labeling budgets
\[
K \in \{ i \cdot C \mid i \in \{1,2,4,8,16,32\} \},
\]
so that the number of labeled instances scales with the number of classes rather than being fixed across datasets. This design enables comparable few-shot regimes across tasks with different output cardinalities, following prior work~\cite{mannix2023cold, dhillonbaseline}. Throughout the paper, we use the notation \(iC\) (\(1C\), \(2C\), \(4C\), etc.) to denote labeling budgets equal to \(i\) times the number of classes in the dataset.

Our main benchmark in Section~\ref{subsec:exp_comparison} evaluates unsupervised selection methods using all 10 official folds per dataset and five random seeds for methods with stochastic components. Results are averaged across seeds and folds for each dataset, and then aggregated across datasets. The supervised genetic search used in Section~\ref{subsec:exp_sensitivity} to estimate the potential gains of optimized selection serves only as an oracle-like diagnostic reference and is therefore restricted to fold 0 due to its computational cost.

The primary evaluation metric is AUC, computed using a one-vs-rest strategy for multiclass datasets, following~\cite{hollmann2025accurate}. Additional ACC and F1 results are reported in Appendix~\ref{sec:app_additional_results}. For method comparison, we report both absolute performance and average ranks across datasets, as rank-based summaries reduce the influence of datasets with substantially different difficulty levels and score ranges. 

For multi-dataset comparison, we perform statistical analyses at the dataset level using fold- and seed-averaged scores for each labeling budget. Following~\cite{demsar2006statistical}, we use Friedman tests over dataset ranks, Nemenyi critical-distance diagrams for post-hoc comparisons, and paired Wilcoxon signed-rank tests against the Random baseline with Benjamini--Hochberg correction applied within each budget~\cite{benjamini1995controlling}. Between-group comparisons use the Mann--Whitney \(U\) (MWU) test~\cite{mann1947test}.

Robustness is further evaluated through leave-one-dataset-out jackknife analysis~\cite{friedl2002jackknife,salinas2024tabrepo}, progressive random dataset removal (10--90\%, 20 orderings), and Spearman correlation between method rank vectors under AUC, ACC, and F1.

\subsection{Baselines}
\label{sec:baselines}

We compare LUCoS against training-free unsupervised selection methods compatible with the same cold-start, label-free setting. For representation-dependent methods such as RDSS and ZCore, we evaluate both the original feature space and the TabClustPFN PIN Encoder embedding space to disentangle the effect of the selection criterion from that of the representation space.

\begin{itemize}
    \item \textbf{Random:} Uniform random sampling from the unlabeled pool.

    \item \textbf{K-Medoids:} Euclidean \(K\)-Medoids in the original feature space, with \(K\) fixed to the labeling budget. Since medoids are real instances, the selected subset can be directly annotated. Implementation details are provided in Appendix~\ref{sec:app_kmedoids_details}.
    
    \item \textbf{ZCore:} An unsupervised scoring-based selection method that favors representative and non-redundant samples through repeated low-dimensional subspace sampling~\cite{griffin2026zero}.

    \item \textbf{RDSS:} An unsupervised subset selection method that iteratively constructs representative and diverse subsets by approximating the global data distribution while discouraging redundancy among selected samples~\cite{shao2024enhancing}.
\end{itemize}

\section{Experiments}
\label{sec:experiments}

Two questions guide this section.

\textbf{1) Does the choice of context instances affect TabPFN's predictions, or does any reasonable subset do equally well?} Section~\ref{subsec:exp_sensitivity} addresses this question through a supervised oracle analysis based on an evolutionary algorithm (SSMA), showing that optimized subset selection improves over random selection by up to $\approx 0.19~\Delta \text{AUC}$ at the smallest labeling budgets, thereby revealing substantial headroom.

\textbf{2) Can that headroom be partially recovered without label access?} The following sections show that LUCoS, K-Medoids applied in the TabClustPFN PIN Encoder embedding space, ranks first across all six labeling budgets under AUC, ACC, and F1, recovering 13--20\% of the oracle gap without labels. Critically, the gains arise from two distinct mechanisms: a selector effect that dominates at 1C, and a representation effect that dominates from 4C onward, particularly rescuing datasets where original-space selection falls below random. Computational details are provided in Appendix~\ref{sec:app_computational_resources}.

\subsection{TabPFN context sensitivity}
\label{subsec:exp_sensitivity}

Since TabPFN performs strongly in low-data regimes, we first evaluate its few-shot behavior. To enable comparisons across datasets with different difficulty levels, we adopt the \textbf{Improvability} metric from prior work~\cite{qu2026tabiclv2}, adapted here to the few-shot setting. Improvability measures the relative AUC achieved by a method under a limited labeling budget with respect to the AUC obtained by TabPFN-2.5 trained on the full labeled training set, which serves as the reference performance.

\begin{figure}[t]
  \centering
  \includegraphics[width=0.4\linewidth]{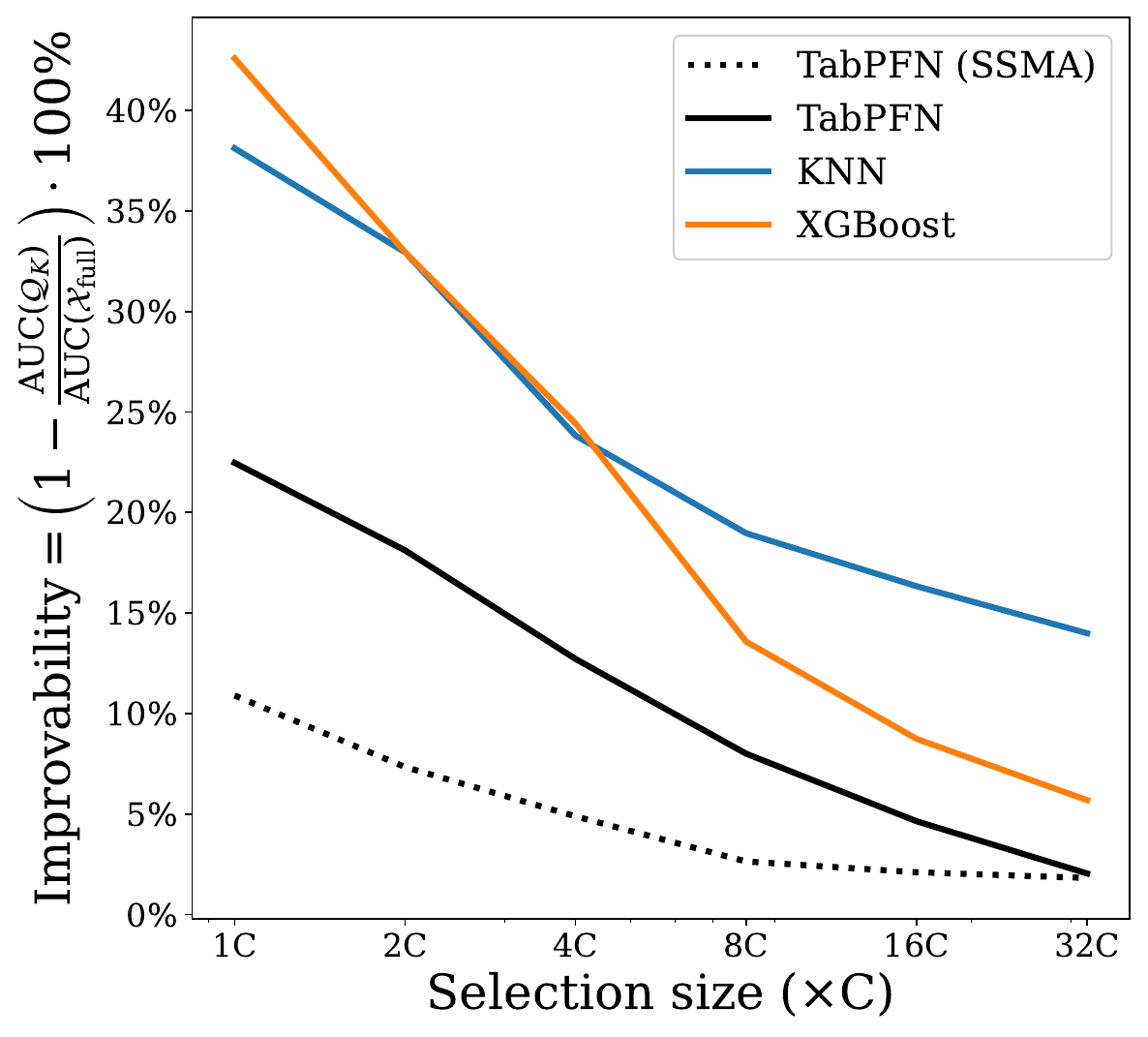}
  \caption{Performance of different classification models versus number of selected training instances under low-label classification (only fold 0). Improvability measures the percentage AUC gap between the selected subset method and the full-data TabPFN reference. Random balanced sampling was used for methods with solid lines. The dotted line shows the performance of TabPFN with the best subset found by SSMA.}
  \label{fig:improvability_vs_num_instances_SSMA}
\end{figure}

Figure~\ref{fig:improvability_vs_num_instances_SSMA} shows that TabPFN consistently outperforms XGBoost~\cite{chen2016xgboost} and KNN baselines under random balanced selection across all labeling budgets. However, different subsets of the same size can induce substantially different predictions, with this sensitivity becoming especially pronounced at very small context sizes. This raises a central question for few-shot tabular learning: given a fixed labeling budget, \textbf{how does predictive performance depend on the selected context instances?}

To quantify this sensitivity, we compare random balanced selection against a supervised oracle-like selection procedure based on the Steady-State Memetic Algorithm (SSMA)~\cite{garcia2008memetic}, chosen for its strong performance in instance selection~\cite{garcia2012prototype}. SSMA explicitly searches for the subset maximizing validation AUC, using label information and validation feedback unavailable in our target cold-start setting. We therefore use it only as an oracle-like reference to estimate the achievable improvement when the context set is directly optimized for TabPFN. Additional implementation details are provided in Appendix~\ref{sec:app_SSMA_implementation}. Due to the high computational cost of this supervised search, the experiment is conducted only on fold 0 of each dataset.

As shown in Figure~\ref{fig:improvability_vs_num_instances_SSMA}, the subsets found by SSMA substantially improve over random selection, particularly at the smallest labeling budgets. The gap reaches approximately \(0.19~\Delta\mathrm{AUC}\) at 1C and narrows steadily as \(K\) increases and more of the training distribution becomes covered. The headroom for selection is therefore largest precisely where labels are scarcest (additional analysis is provided in Appendix~\ref{subsec:apx_oracle}). These findings motivate the main question addressed in this work: if TabPFN is both highly effective in few-shot regimes and highly sensitive to subset composition, \textbf{can context selection be performed without labels by exploiting unsupervised structure in tabular representation spaces?}

\subsection{Benchmark comparison}
\label{subsec:exp_comparison}

We compare LUCoS against random selection, original-space K-Medoids, and two recent unsupervised baselines (RDSS and ZCore) applied in both representation spaces. Table~\ref{tab:auc_and_auc_rank} reports mean AUC and mean AUC rank across the 67 datasets at every budget. LUCoS obtains the best mean rank at every budget, while RDSS and ZCore fail to reliably beat random selection in either space.

\begin{table}[b]
\centering
\scriptsize
\setlength{\tabcolsep}{3pt}
\begin{tabular}{ll *{6}{cc}}
\toprule
\multirow{2}{*}{Space} & \multirow{2}{*}{Method}
  & \multicolumn{2}{c}{1C}
  & \multicolumn{2}{c}{2C}
  & \multicolumn{2}{c}{4C}
  & \multicolumn{2}{c}{8C}
  & \multicolumn{2}{c}{16C}
  & \multicolumn{2}{c}{32C} \\
\cmidrule(lr){3-4}\cmidrule(lr){5-6}\cmidrule(lr){7-8}%
\cmidrule(lr){9-10}\cmidrule(lr){11-12}\cmidrule(lr){13-14}
& & $\text{AUC}{\uparrow}$ & $\text{R}_{\text{AUC}}{\downarrow}$
  & AUC & $\text{R}_{\text{AUC}}$
  & AUC & $\text{R}_{\text{AUC}}$
  & AUC & $\text{R}_{\text{AUC}}$
  & AUC & $\text{R}_{\text{AUC}}$
  & AUC & $\text{R}_{\text{AUC}}$ \\
\midrule
-- & \textit{Random Unsupervised}
  & 0.61            & \textit{3.45}
  & 0.68            & \textit{3.36}
  & 0.74            & \textit{3.34}
  & \textbf{0.80}   & \textit{3.18}
  & 0.83            & \textit{3.24}
  & \textbf{0.86}   & \textit{3.34} \\
\midrule
\multirow{3}{*}{Original}
& RDSS
  & 0.60 & 4.26
  & 0.65 & 4.15$^{\circ}$
  & 0.70 & 4.10$^{\dagger}$
  & 0.75 & 4.07$^{\ddagger}$
  & 0.79 & 4.40$^{\dagger}$
  & 0.83 & 4.38$^{\ddagger}$ \\
& ZCore
  & 0.54 & 5.49$^{\ddagger}$
  & 0.59 & 5.70$^{\ddagger}$
  & 0.64 & 5.68$^{\ddagger}$
  & 0.71 & 5.34$^{\ddagger}$
  & 0.78 & 5.28$^{\ddagger}$
  & 0.82 & 5.05$^{\ddagger}$ \\
& K-Medoids
  & \textbf{0.65} & \textit{2.77}$^{\ddagger}$
  & 0.69          & \textit{3.13}
  & 0.74          & \textit{3.13}
  & 0.78          & \textit{3.48}
  & 0.82          & \textit{3.43}
  & 0.85          & \textit{3.37} \\
\midrule
\multirow{3}{*}{\makecell{TabClustPFN \\ PIN Encoder}}
  & RDSS
  & 0.58 & 4.76$^{\ddagger}$
  & 0.62 & 4.52$^{\ddagger}$
  & 0.68 & 4.93$^{\ddagger}$
  & 0.73 & 4.98$^{\ddagger}$
  & 0.80 & 4.76$^{\ddagger}$
  & 0.84 & 4.75$^{\ddagger}$ \\
& ZCore
  & 0.57 & 4.67$^{\ddagger}$
  & 0.63 & 4.62$^{\ddagger}$
  & 0.71 & 4.28$^{\dagger}$
  & 0.77 & 4.33$^{\ddagger}$
  & 0.82 & 4.13$^{\circ}$
  & 0.85 & \textit{4.01} \\
& \textbf{K-Medoids (LUCoS)}
  & \textbf{0.65} & \textbf{2.60}$^{\ddagger}$
  & \textbf{0.71} & \textbf{2.52}$^{\ddagger}$
  & \textbf{0.76} & \textbf{2.54}$^{\dagger}$
  & \textbf{0.80} & \textbf{2.63}
  & \textbf{0.84} & \textbf{2.75}
  & \textbf{0.86} & \textbf{3.09} \\
\bottomrule
\end{tabular}
\vspace{5pt}
\caption{%
  Mean AUC and mean rank $\mathrm{R}_{\mathrm{AUC}}$ per labeling budget $K$
  (67 datasets $\times$ 10 folds).
  \textbf{LUCoS} is K-Medoids in the TabClustPFN PIN Encoder embedding space.   
  \textbf{Bold}: best per column (rounded to 2\,d.p.).
  \textit{Italic}: within the Nemenyi critical-distance band of the best-ranked
  method (Friedman $p \leq 5{\times}10^{-9}$ at every budget;
  $\mathrm{CD}{=}1.10$).
  Superscripts: Wilcoxon vs.\ Random Unsupervised, Benjamini--Hochberg correction;
  $^{\circ}$\,$q{<}0.05$,\;
  $^{\dagger}$\,$q{<}0.01$,\;
  $^{\ddagger}$\,$q{<}0.001$.
  Rankings under ACC and F1 are highly correlated with AUC
  (Spearman $\rho{=}0.87$ and $\rho{=}0.95$).
  Full results in Appendix~\ref{sec:app_additional_results}.
}
\label{tab:auc_and_auc_rank}
\end{table}

\begin{figure}[t]
    \centering
    \includegraphics[width=0.4\linewidth]{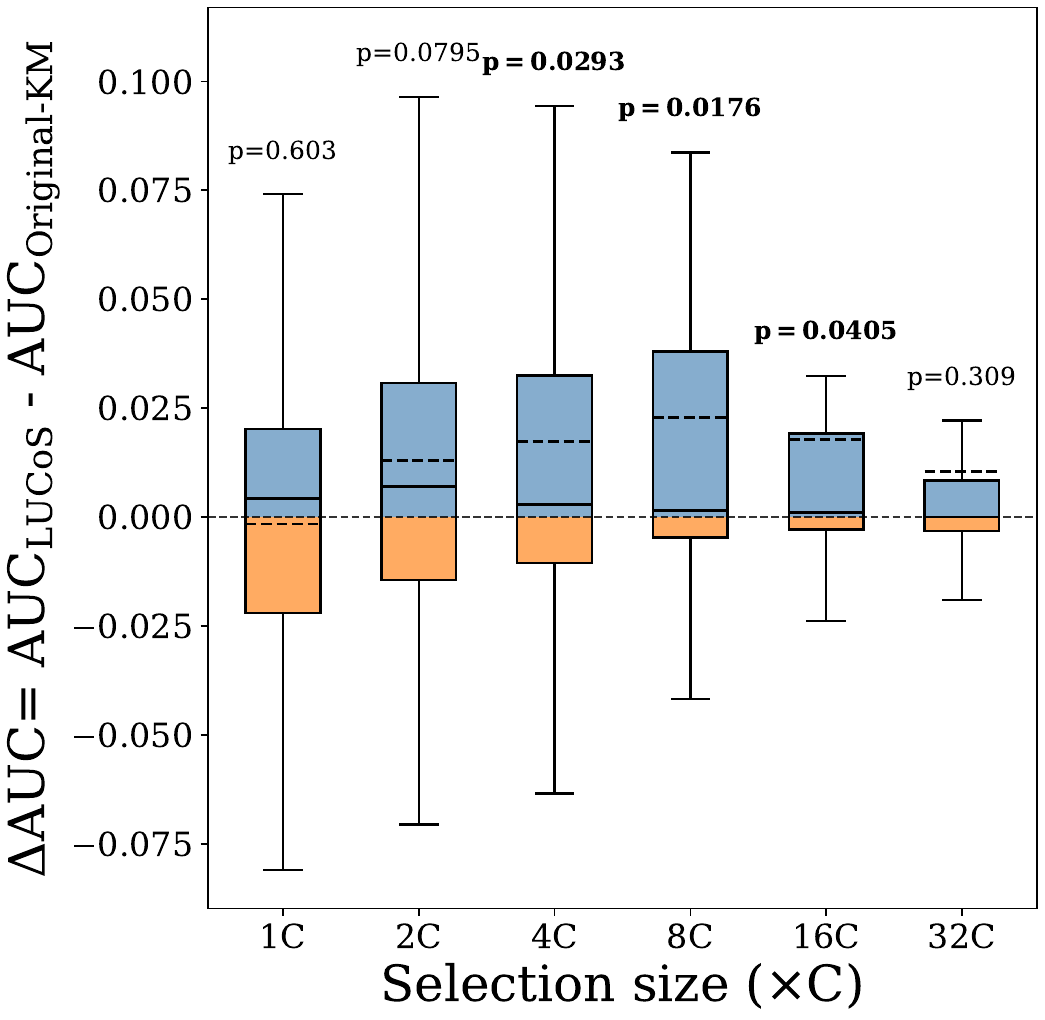}
    \caption{Paired AUC difference
    $\Delta\text{AUC} = \text{AUC}_{\text{LUCoS}} - \text{AUC}_{\text{Original-KM}}$
    per labeling budget ($n = 67$ datasets).
    Boxes are split at zero (blue: LUCoS advantage; orange: Original-KM advantage);
    solid and dashed lines mark the median and mean.
    Wilcoxon signed-rank $p$-values are annotated above each boxplot
    (bold: $p < 0.05$).
    The advantage is near zero and non-significant at 1C--2C,
    shifts positively and reaches significance at 4C--16C,
    then fades at 32C.}
    \label{fig:lucosboxplot}
\end{figure}

The most informative comparison is LUCoS versus original-space K-Medoids, because the selector is fixed and only the representation changes. The advantage is not uniform but follows a clear pattern: the two methods are statistically tied at 1C--2C, LUCoS separates significantly at 4C--16C, and the gap narrows again at 32C as all methods saturate (Figure~\ref{fig:lucosboxplot}). This pattern is non-trivial: Original-space $K$-Medoids degrades steadily as $K$ grows, falling below Random Unsupervised in mean AUC from 8C onward, while LUCoS holds the top rank throughout. Holding the selector fixed and switching only the representation thus converts a method that fails at practical budgets into one that is best-ranked across budgets.

The TabClustPFN representation does not rescue other selectors: RDSS and ZCore remain below Random Unsupervised across budgets in both spaces. The improvement is therefore not attributable to either ingredient alone but to their combination, the geometry of TabClustPFN paired with K-Medoids. Section~\ref{subsec:exp_representation} decomposes this pattern to explain both the 1C tie and the mid-budget divergence.

\subsection{Why representation matters}
\label{subsec:exp_representation}

We diagnose the source of LUCoS's advantage by decomposing its gain over Random Unsupervised into two components:
\[
\Delta_{\mathrm{Sel}} = \mathrm{AUC}(\text{Original-KM}) - \mathrm{AUC}(\text{Random}), \qquad
\Delta_{\mathrm{Repr}} = \mathrm{AUC}(\text{LUCoS}) - \mathrm{AUC}(\text{Original-KM}),
\]
where Original-KM denotes K-Medoids applied in the original feature space, and significance against zero is assessed by one-sample Wilcoxon signed-rank tests.

\begin{figure}[t]
    \centering
    \begin{subfigure}[t]{0.49\linewidth}
        \centering
        \includegraphics[width=\linewidth]{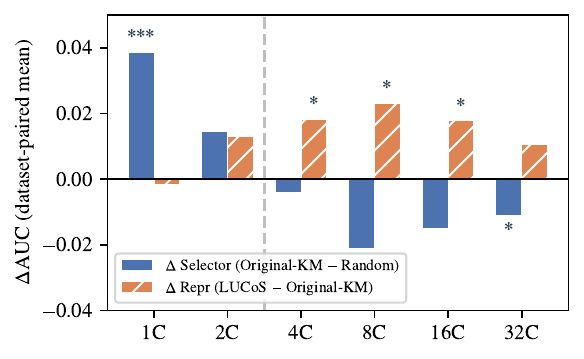}
        \caption{Selector vs.\ representation decomposition}
        \label{fig:gain_decomposition_a}
    \end{subfigure}
    \hfill
    \begin{subfigure}[t]{0.49\linewidth}
        \centering
        \includegraphics[width=\linewidth]{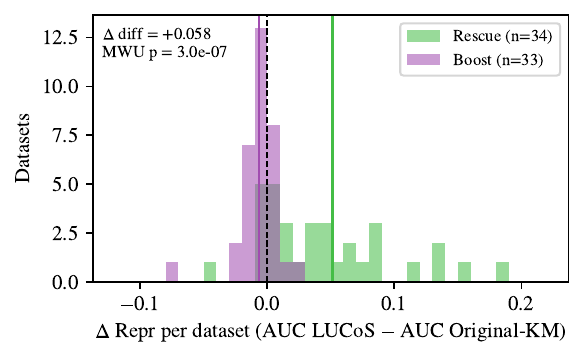}
        \caption{Rescue vs.\ Boost distribution of $\Delta_\text{Repr}$ at 8C}
        \label{fig:gain_decomposition_b}
    \end{subfigure}
    \caption{%
        \textbf{Left:} Decomposition of the LUCoS gain over \textit{Random Unsupervised}
        into $\Delta_\text{Sel}$ (blue bars) and $\Delta_\text{Repr}$ (orange hatched bars),
        shown as dataset-paired means per budget.
        Stars indicate Wilcoxon significance against zero:
        $^{*}p{<}0.05$,\; $^{**}p{<}0.01$,\; $^{***}p{<}0.001$.
        The dashed vertical line marks the 2C--4C crossover, beyond which $\Delta_\text{Sel}$
        is non-significant and $\Delta_\text{Repr}$ becomes the sole significant component.
        \textbf{Right:} Distribution of $\Delta_\text{Repr}$ at 8C (the budget with the
        largest group separation), partitioned into Rescue ($n{=}34$, orange) and
        Boost ($n{=}33$, blue) groups.
        Rescue datasets gain a mean $\Delta_\text{Repr} = {+}0.058$;
        Boost datasets show near-zero average change.
        Mann--Whitney $U$: $p = 3{\times}10^{-7}$.
    }
    \label{fig:gain_decomposition}
\end{figure}

Figure~\ref{fig:gain_decomposition_a} reveals a polarity reversal between 1C and 4C. At 1C the selector drives nearly the entire gain ($\Delta_\text{Sel}$ significant, $\Delta_\text{Repr}$ near zero): K-Medoids on raw features already captures the available structure, and the latent space contributes nothing further. From 4C onward the picture inverts. The selector component becomes negligible and eventually negative at the largest budgets, while the representation component is the only positive and significant contributor through 16C, fading at 32C as methods saturate. At 1C the selector alone recovers ${\approx}20\%$ of the supervised oracle gap from Section~\ref{subsec:exp_sensitivity}; from 4C to 8C the representation alone recovers ${\approx}16\%\text{--}30\%$ of that gap.

The average representation gain at 4C--8C is modest, but this average hides a bimodal structure. Partitioning datasets at each budget into a \emph{Rescue} group (Original-KM below Random) and a \emph{Boost} group (Original-KM above Random) shows a striking contrast: at 8C, Rescue datasets ($n=34$) gain a mean $\Delta_\text{Repr} = +0.058$ from the TabClustPFN representation, while Boost datasets ($n=33$) show essentially no change (Figure~\ref{fig:gain_decomposition_b}; Mann--Whitney $U$ (MWU), $p = 3{\times}10^{-7}$). The same bimodal pattern holds at every budget (Appendix~\ref{subsec:apx_rescue}). The latent space therefore does two distinct things: it actively compensates where original-space selection fails, and it imposes no cost where it does not. The Rescue regime is also the dominant scenario: 52 of 67 datasets (78\%) experience Original-KM falling below random at some budget. LUCoS is therefore the only method that performs well in both regimes: at 1C through the K-Medoids coverage effect, and from 4C onward through the latent representation, where Original-KM falls below random on most datasets.

\subsection{Robustness and scope}
\label{subsec:exp_robustness}

LUCoS's top AUC rank is stable under the reported robustness checks. Under leave-one-dataset-out, removing each of the 67 datasets in turn produces zero changes to the top-ranked method across all six budgets (402 configurations), with a maximum rank shift of 0.07. Under progressive random removal, LUCoS retains the top rank in a majority of 20 random orderings through 60\% dataset removal; beyond that point the surviving benchmark is too small for rank averaging to stabilize. The ranking also transfers across metrics: LUCoS ranks first under ACC and F1 as well as under AUC. The picture across all three checks is consistent — LUCoS's lead is not driven by any individual dataset and is not an artifact of the evaluation metric.

\section{Conclusions, limitations, and future work} \label{sec:conclusions_limitations}

We studied unsupervised context selection for TFMs — the problem of choosing which instances to annotate before any labels exist. Our experiments show, first, that few-shot TabPFN is highly sensitive to the specific subset of labeled instances used as context, with a supervised oracle exposing substantial headroom that better selection could in principle recover. Second, they show that this headroom is not closed by transferring distance- or coverage-based selection methods designed for other modalities: RDSS and ZCore fail to beat random sampling reliably, and even simple K-Medoids in the original feature space falls below random on the majority of datasets as the budget grows.

The remedy is not a more sophisticated selection method but a better geometry. LUCoS applies K-Medoids in the latent space of an unsupervised tabular PFN and is the top-ranked method at every labeling budget under AUC, ACC, and F1. A gain decomposition reveals a clean mechanism: at the smallest budgets the gain comes from enforcing coverage rather than sampling randomly, and the latent space contributes nothing beyond what raw features already provide; from mid-range budgets onward, raw-space coverage degrades performance on most datasets, and the latent space is what restores reliable behavior. The two effects are complementary, and LUCoS alone captures both.

Taken together, these findings reframe unsupervised context selection in tabular few-shot learning: reliable improvement depends less on the sophistication of the selection method than on the geometry in which representativeness is defined. Improving representation learning for tabular data may therefore be as important as designing increasingly complex selection heuristics, positioning unsupervised context selection as a natural bridge between abundant unlabeled data and the strong but context-sensitive predictions of TFMs.

\paragraph{Limitations and future work.}

Several limitations bound the scope of our claims and indicate directions for future work. First, our benchmark inherits the constraints of TabPFN-2.5: we evaluate on classification datasets with at most ten classes, and because labeling budgets scale with the number of classes, conclusions for higher-cardinality problems remain open. Second, we evaluate exclusively on classification; extending LUCoS to regression is an important next step, where the notion of class coverage no longer applies and subset quality may depend more strongly on the target function's geometry. Third, K-Medoids becomes computationally expensive on very large unlabeled pools, and scalable approximations or streaming variants would be needed for deployment at industrial scale. 

LUCoS also leaves open several method-specific questions. It does not explicitly enforce class- or minority-region coverage. While our results suggest robustness in many imbalanced settings, severe imbalance remains an open challenge in the cold-start regime, where labels are unavailable to inform stratified selection. Finally, LUCoS is currently coupled to TabClustPFN as its representation source. Whether the observed gains generalize to other tabular embedding models, whether representations can be trained explicitly for instance selection rather than clustering alone, and whether alternative geometry-based selection strategies outperform K-Medoids within the same latent framework remain open questions. Beyond predictive performance, the selected medoids may also serve as compact dataset summaries, suggesting a link between context selection and dataset-level explainability.

\section{Acknowledgments}

This publication is part of the Project ``Ethical, Responsible and General Purpose Artificial Intelligence: Applications in Risk Scenarios'' (IAFER) Exp.:TSI-100927-2023-1 funded through the Creation of university-industry research programs (ENIA Programs), aimed at the research and development of artificial intelligence, for its dissemination and education within the framework of the Recovery, Transformation and Resilience Plan from the European Union Next Generation EU through the Ministry for Digital Transformation and the Civil Service. This work is also supported by the Knowledge Generation Project PID2023-149128NB-I00, funded by MICIU/AEI/10.13039/501100011033 and by ERDF, EU.

R. Romero-Zaliz is also funded by Consejería de Universidad, Investigación e Innovación de la Junta de Andalucía (DGP\_PIDI\_2024\_00563).

\bibliographystyle{ieeetr}
\bibliography{bibliography}

\newpage

\appendix

\section{Societal Impact Statement} \label{sec:app_social_impact}

We do not identify significant foreseeable risks associated with this work beyond those already present in existing unsupervised instance selection methods. Our goal is to study instance selection strategies for low-label tabular learning, and our approach operates purely at the data selection stage without introducing new modeling biases or decision-making criteria. Furthermore, our method improves data efficiency by reducing the number of labeled instances required to achieve strong predictive performance. This can substantially decrease the time, cost, and human effort involved in annotation, particularly in domains where labeling requires expert knowledge or costly procedures. In some applications, reducing labeling needs may also mitigate indirect negative effects associated with data collection, such as environmental impact or health risks when annotations depend on physical experiments. Moreover, by highlighting the strong few-shot capabilities of Prior-Fitted Networks such as TabPFN, our work encourages further research into instance selection methods tailored to these models, potentially leading to more efficient and scalable classification pipelines. Overall, we view our contribution as a general-purpose, low-risk improvement that can facilitate more accessible, cost-effective, and resource-efficient deployment of tabular machine learning systems.

\section{Computational resources} \label{sec:app_computational_resources}
Experiments were conducted on a machine with an AMD EPYC 7742 CPU, 128 GB RAM, and an NVIDIA Tesla A100 GPU (40 GB VRAM). Selection methods were executed on the CPU, while TabPFN inference was performed on the GPU. For larger datasets, memory usage approached the available capacity, whereas smaller datasets were processed well within these limits.

\section{K-Medoids implementation details} \label{sec:app_kmedoids_details}

We implement K-medoids using the \textit{sklearn\_extra} library~\footnote{\url{https://scikit-learn-extra.readthedocs.io/en/stable/generated/sklearn_extra.cluster.KMedoids.html}. Last accessed \today}. Categorical features are one-hot encoded to avoid introducing spurious ordinal relationships and to enable standard distance metrics.

We use the \texttt{kmedoids++} initialization strategy, which adapts the KMeans++ seeding procedure \cite{arthur2007k} by selecting the first medoid uniformly at random and sampling subsequent medoids with probability proportional to their squared distance from the nearest selected medoid. This encourages a diverse initialization and improves stability. All remaining hyperparameters are set to their default values: \texttt{metric='euclidean'}, \texttt{method='alternate'}, and \texttt{max\_iter=300}.

\section{Additional Results} 
\label{sec:app_additional_results}

\subsection{SSMA oracle gap}
\label{subsec:apx_oracle}

Table~\ref{tab:ssma_oracle_appendix} presents the data behind
Figure~\ref{fig:improvability_vs_num_instances_SSMA} and the supervised-oracle
motivation in Section~\ref{subsec:exp_sensitivity}. SSMA is a memetic
supervised search for the context subset of size $K$ that maximizes validation
AUC; we use it as a diagnostic upper bound on what context selection could in
principle achieve, not as a deployable method. Because the search is
computationally prohibitive, results are restricted to fold 0 of every dataset.
The gap between SSMA and random selection narrows monotonically with $K$, from
$+0.185~\Delta\mathrm{AUC}$ at 1C to $+0.017~\Delta\mathrm{AUC}$ at 32C,
confirming that the headroom for unsupervised selection is largest precisely
where labels are scarcest --- the regime where any practical method has to
operate. The dataset-paired median lies consistently below the mean, indicating
a long tail of datasets where the achievable improvement is substantially
larger than the average suggests.

\begin{table}[h]
\centering
\caption{Supervised-oracle headroom on the OpenML-CC18 benchmark (fold 0). For
each labeling budget, SSMA searches for the context subset of size $K = c\cdot C$
that maximizes validation AUC; Random Unsupervised samples uniformly.
$\overline{\Delta}$ is the dataset-paired mean improvement of SSMA over Random,
$\widetilde{\Delta}$ the corresponding median. Restricted to fold 0 due to the
cost of the SSMA search}
\label{tab:ssma_oracle_appendix}
\setlength{\tabcolsep}{5pt}
\begin{tabular}{lccccc}
\toprule
Budget & $n_{\mathrm{datasets}}$ & SSMA $\overline{\mathrm{AUC}}$
       & Random $\overline{\mathrm{AUC}}$
       & $\overline{\Delta}$ & $\widetilde{\Delta}$ \\
\midrule
1C  & 67 & 0.7934 & 0.6080 & $+0.1854$ & $+0.1740$ \\
2C  & 67 & 0.8262 & 0.6772 & $+0.1491$ & $+0.1208$ \\
4C  & 67 & 0.8473 & 0.7382 & $+0.1091$ & $+0.1008$ \\
8C  & 67 & 0.8663 & 0.7955 & $+0.0708$ & $+0.0575$ \\
16C & 67 & 0.8748 & 0.8372 & $+0.0376$ & $+0.0181$ \\
32C & 67 & 0.8804 & 0.8634 & $+0.0169$ & $+0.0024$ \\
\bottomrule
\end{tabular}
\end{table}

\subsection{Method comparison under ACC and F1}
\label{subsec:apx_methods}

Tables~\ref{tab:acc_ranks_appendix} and~\ref{tab:f1_ranks_appendix} expand the
AUC analysis of Section~\ref{subsec:exp_comparison} to accuracy and macro-F1.
Mean values and mean ranks are computed identically to Table~\ref{tab:auc_and_auc_rank}
(67 datasets $\times$ 10 folds; same seven methods; same six budgets). LUCoS
retains the best mean rank at most budgets under both metrics, confirming that
the ordering observed under AUC is not metric-specific. Method-rank vectors
between metrics correlate at Spearman $\rho = 0.87$ (AUC vs.\ ACC) and
$\rho = 0.95$ (AUC vs.\ F1), averaged across budgets.

\begin{table}[h]
\centering
\caption{%
 Mean ACC and mean rank $\mathrm{R}_{\mathrm{ACC}}$ across 67 datasets~$\times$~10
 folds per labeling budget $K$ (higher ACC is better; lower rank is better).
 \textbf{LUCoS} is K-Medoids in the TabClustPFN PIN Encoder embedding space.   
 \textbf{Bold} marks the best value per column (ties included).
 \textit{Italic} marks rank values within the Nemenyi critical-distance band of the
 best method at that budget (Friedman $\chi^{2}$ significant at every budget,
 $p \leq 9{\times}10^{-9}$; critical distance $\mathrm{CD}{=}1.10$, $\alpha{=}0.05$, $k{=}7$, $n{=}67$  ).
 Superscripts denote significant paired differences against Random Unsupervised
 after Benjamini--Hochberg correction; direction is given by the corresponding $\mathrm{R}_{\mathrm{ACC}}$:
 $^{\circ}$\,$q{<}0.05$,\; $^{\dagger}$\,$q{<}0.01$,\; $^{\ddagger}$\,$q{<}0.001$.
}
\label{tab:acc_and_acc_rank}
\scriptsize
\setlength{\tabcolsep}{3pt}
\begin{tabular}{ll *{6}{cc}}
\toprule
\multirow{2}{*}{Space} & \multirow{2}{*}{Method}
 & \multicolumn{2}{c}{1C}
 & \multicolumn{2}{c}{2C}
 & \multicolumn{2}{c}{4C}
 & \multicolumn{2}{c}{8C}
 & \multicolumn{2}{c}{16C}
 & \multicolumn{2}{c}{32C}
 \\
\cmidrule(lr){3-4}\cmidrule(lr){5-6}\cmidrule(lr){7-8}%
\cmidrule(lr){9-10}\cmidrule(lr){11-12}\cmidrule(lr){13-14}
& & $\text{ACC}{\uparrow}$ & $\mathrm{R}_{\mathrm{ACC}}{\downarrow}$ & ACC & $\mathrm{R}_{\mathrm{ACC}}$ & ACC & $\mathrm{R}_{\mathrm{ACC}}$ & ACC & $\mathrm{R}_{\mathrm{ACC}}$ & ACC & $\mathrm{R}_{\mathrm{ACC}}$ & ACC & $\mathrm{R}_{\mathrm{ACC}}$ \\
\midrule
-- & \textit{Random Unsupervised} & 0.54 & 4.21 & 0.60 & 4.09 & 0.67 & 3.98 & 0.73 & \textit{3.66} & \textbf{0.78} & \textit{3.54} & \textbf{0.81} & \textit{3.33} \\
\midrule
\multirow{3}{*}{Original} & RDSS & 0.54 & \textit{3.86} & 0.59 & \textit{3.60} & 0.66 & \textit{3.55} & 0.71 & \textit{3.68} & 0.74 & \textit{3.75} & 0.77 & \textit{3.81}$^{\circ}$ \\
 & ZCore & 0.46 & 4.75$^{\dagger}$ & 0.50 & 4.84$^{\ddagger}$ & 0.56 & 4.93$^{\ddagger}$ & 0.65 & 4.84$^{\ddagger}$ & 0.71 & 5.10$^{\ddagger}$ & 0.75 & 5.28$^{\ddagger}$ \\
 & K-Medoids & 0.58 & \textbf{3.05}$^{\ddagger}$ & 0.62 & \textit{3.30}$^{\circ}$ & 0.68 & \textit{3.49}$^{\circ}$ & 0.73 & \textit{3.30} & 0.77 & \textit{3.36} & 0.80 & \textit{3.54} \\
\midrule
\multirow{3}{*}{\makecell{TabClustPFN \\ PIN Encoder}}
 & RDSS & 0.52 & \textit{3.99} & 0.56 & 4.09 & 0.62 & 4.25$^{\dagger}$ & 0.67 & 4.59$^{\ddagger}$ & 0.73 & 4.75$^{\ddagger}$ & 0.77 & 4.61$^{\ddagger}$ \\
 & ZCore & 0.48 & 5.01$^{\ddagger}$ & 0.52 & 5.24$^{\ddagger}$ & 0.60 & 5.07$^{\ddagger}$ & 0.67 & 4.93$^{\ddagger}$ & 0.74 & 4.60$^{\ddagger}$ & 0.78 & 4.33$^{\dagger}$ \\
 & K-Medoids (LUCoS) & \textbf{0.59} & \textit{3.13}$^{\ddagger}$ & \textbf{0.64} & \textbf{2.84}$^{\ddagger}$ & \textbf{0.70} & \textbf{2.74}$^{\ddagger}$ & \textbf{0.74} & \textbf{3.00}$^{\dagger}$ & \textbf{0.78} & \textbf{2.90}$^{\circ}$ & \textbf{0.81} & \textbf{3.10} \\
\bottomrule
\end{tabular}
\label{tab:acc_ranks_appendix}
\end{table}

\begin{table}[h!]
\centering
\caption{%
 Mean F1 and mean rank $\mathrm{R}_{\mathrm{F1}}$ across 67 datasets~$\times$~10
 folds per labeling budget $K$ (higher F1 is better; lower rank is better).
 \textbf{LUCoS} is K-Medoids in the TabClustPFN PIN Encoder embedding space.   
 \textbf{Bold} marks the best value per column (ties included).
 \textit{Italic} marks rank values within the Nemenyi critical-distance band of the
 best method at that budget (Friedman $\chi^{2}$ significant at every budget,
 $p \leq 3{\times}10^{-11}$; critical distance $\mathrm{CD}{=}1.10$, $\alpha{=}0.05$, $k{=}7$, $n{=}67$).
 Superscripts denote significant paired differences against Random Unsupervised
 after Benjamini--Hochberg correction; direction is given by the corresponding $\mathrm{R}_{\mathrm{F1}}$:
 $^{\circ}$\,$q{<}0.05$,\; $^{\dagger}$\,$q{<}0.01$,\; $^{\ddagger}$\,$q{<}0.001$.
}
\label{tab:f1_and_f1_rank}
\scriptsize
\setlength{\tabcolsep}{3pt}
\begin{tabular}{ll *{6}{cc}}
\toprule
\multirow{2}{*}{Space} & \multirow{2}{*}{Method}
 & \multicolumn{2}{c}{1C}
 & \multicolumn{2}{c}{2C}
 & \multicolumn{2}{c}{4C}
 & \multicolumn{2}{c}{8C}
 & \multicolumn{2}{c}{16C}
 & \multicolumn{2}{c}{32C}
 \\
\cmidrule(lr){3-4}\cmidrule(lr){5-6}\cmidrule(lr){7-8}%
\cmidrule(lr){9-10}\cmidrule(lr){11-12}\cmidrule(lr){13-14}
& & $\text{F}_1{\uparrow}$ & $\mathrm{R}_{\mathrm{F}_1}{\downarrow}$ & F$_1$ & $\mathrm{R}_{\mathrm{F}_1}$ & F$_1$ & $\mathrm{R}_{\mathrm{F}_1}$ & F$_1$ & $\mathrm{R}_{\mathrm{F}_1}$ & F$_1$ & $\mathrm{R}_{\mathrm{F}_1}$ & F$_1$ & $\mathrm{R}_{\mathrm{F}_1}$ \\
\midrule
-- & \textit{Random Unsupervised} & 0.37 & 3.64 & 0.45 & \textit{3.55} & 0.53 & \textit{3.41} & 0.61 & \textit{3.22} & \textbf{0.67} & \textit{3.22} & \textbf{0.71} & \textit{3.31} \\
\midrule
\multirow{3}{*}{Original} & RDSS & 0.37 & 3.96 & 0.43 & 3.72 & 0.51 & \textit{3.72} & 0.57 & \textit{3.62} & 0.61 & \textit{3.81}$^{\circ}$ & 0.66 & \textit{3.76}$^{\circ}$ \\
 & ZCore & 0.27 & 5.66$^{\ddagger}$ & 0.31 & 5.99$^{\ddagger}$ & 0.37 & 6.08$^{\ddagger}$ & 0.48 & 5.76$^{\ddagger}$ & 0.56 & 5.44$^{\ddagger}$ & 0.62 & 5.31$^{\ddagger}$ \\
 & K-Medoids & 0.42 & \textit{2.81}$^{\dagger}$ & 0.48 & \textit{2.84}$^{\dagger}$ & 0.55 & \textbf{2.67}$^{\circ}$ & 0.61 & \textit{3.14} & 0.66 & \textit{3.23} & 0.70 & \textit{3.27} \\
\midrule
\multirow{3}{*}{\makecell{TabClustPFN \\ PIN Encoder}} & RDSS & 0.33 & 4.63$^{\dagger}$ & 0.38 & 4.54$^{\ddagger}$ & 0.46 & 4.66$^{\ddagger}$ & 0.52 & 4.91$^{\ddagger}$ & 0.61 & 4.57$^{\ddagger}$ & 0.67 & 4.75$^{\ddagger}$ \\
 & ZCore & 0.31 & 4.93$^{\ddagger}$ & 0.36 & 4.91$^{\ddagger}$ & 0.44 & 4.75$^{\ddagger}$ & 0.53 & 4.73$^{\ddagger}$ & 0.61 & 4.80$^{\ddagger}$ & 0.68 & 4.43$^{\ddagger}$ \\
 & K-Medoids (LUCoS) & \textbf{0.43} & \textbf{2.37}$^{\ddagger}$ & \textbf{0.49} & \textbf{2.46}$^{\ddagger}$ & \textbf{0.56} & \textit{2.69}$^{\ddagger}$ & \textbf{0.62} & \textbf{2.61}$^{\circ}$ & \textbf{0.67} & \textbf{2.93} & \textbf{0.71} & \textbf{3.16} \\
\bottomrule
\end{tabular}
\label{tab:f1_ranks_appendix}
\end{table}

\subsection{Leave-one-dataset-out and progressive removal}
\label{subsec:apx_lodo}

We expand the two dataset-level robustness checks introduced in
Section~\ref{subsec:exp_robustness}. Across all 402 leave-one-dataset-out
configurations (67 datasets $\times$ 6 budgets), LUCoS is never displaced from
the top rank, and the maximum absolute shift in its mean rank caused by removing
any single dataset stays below $0.07$ at every budget
(Table~\ref{tab:lodo_appendix}). Under progressive random removal
(Table~\ref{tab:progressive_appendix}), LUCoS retains the top rank in a
majority of 20 random orderings through 60\% dataset removal; the majority is
lost at 70\%, by which point fewer than twenty datasets remain and rank
averaging becomes unstable.

\begin{table}[h!]
\centering
\caption{Leave-one-dataset-out diagnostic. For each labeling budget, each of
the 67 datasets is removed in turn and the mean-rank table is recomputed.
\textit{Lost top} counts configurations in which LUCoS no longer holds the
best mean rank; \textit{Max rank shift} is the largest absolute change in
LUCoS's mean rank across the 67 deletions.}
\label{tab:lodo_appendix}
\begin{tabular}{lccc}
\toprule
Budget & Configurations & Lost top & Max rank shift \\
\midrule
1C  & 67 & 0 & 0.067 \\
2C  & 67 & 0 & 0.068 \\
4C  & 67 & 0 & 0.068 \\
8C  & 67 & 0 & 0.066 \\
16C & 67 & 0 & 0.049 \\
32C & 67 & 0 & 0.059 \\
\bottomrule
\end{tabular}
\end{table}

\begin{table}[h]
\centering
\caption{Progressive random removal of datasets. For each removal fraction, 20
random orderings are drawn; \textit{Seeds with LUCoS top} counts the orderings in which LUCoS retains the best mean rank simultaneously at every budget.
\textit{Datasets kept} is the size of the surviving benchmark.}
\label{tab:progressive_appendix}
\begin{tabular}{lccc}
\toprule
Fraction removed & Datasets kept & Seeds with LUCoS top & Total seeds \\
\midrule
0.1 & 60 & 19 & 20 \\
0.2 & 53 & 17 & 20 \\
0.3 & 46 & 15 & 20 \\
0.4 & 40 & 15 & 20 \\
0.5 & 33 & 12 & 20 \\
0.6 & 26 & 11 & 20 \\
0.7 & 20 &  8 & 20 \\
0.8 & 13 &  6 & 20 \\
0.9 &  6 &  2 & 20 \\
\bottomrule
\end{tabular}
\end{table}

\subsection{Rescue and Boost partition across budgets}
\label{subsec:apx_rescue}

Table~\ref{tab:rescue_boost_appendix} expands the Rescue/Boost partition shown
for 8C in Figure~\ref{fig:gain_decomposition_b} to all six labeling budgets.
At every budget, datasets where original-space K-Medoids falls below Random
(Rescue group) gain substantially from the TabClustPFN representation, while
datasets where it does not (Boost group) show essentially no change. The
between-group separation is highly significant throughout (Mann--Whitney $U$,
$p \leq 10^{-3}$), and the dominance of Rescue over Boost in mean
$\Delta_{\mathrm{Repr}}$ is stable at $\approx 0.05$ across 1C--16C, halving
at 32C as all methods saturate. The size of the Rescue group also grows with
budget — from 22 datasets at 1C to 35 at 32C — consistent with the main-text
observation that original-space selection degrades as more instances are
labeled.

\begin{table}[h]
\centering
\caption{Rescue vs.\ Boost decomposition of $\Delta_{\mathrm{Repr}}$ across all
labeling budgets. Datasets are partitioned at each budget into Rescue
(Original-space K-Medoids falls strictly below Random Unsupervised) and Boost
(otherwise); the headline 8C row corresponds to
Figure~\ref{fig:gain_decomposition_b}. \textit{$\Delta\Delta$} is the
between-group difference of means; \textit{MWU $p$} is the Mann--Whitney
two-sided $p$-value.}
\label{tab:rescue_boost_appendix}
\setlength{\tabcolsep}{5pt}
\begin{tabular}{lcccccc}
\toprule
Budget & $n_{\mathrm{Rescue}}$ & $\overline{\Delta}_{\mathrm{Repr}}^{\mathrm{Rescue}}$
       & $n_{\mathrm{Boost}}$  & $\overline{\Delta}_{\mathrm{Repr}}^{\mathrm{Boost}}$
       & $\Delta\Delta$ & MWU $p$ \\
\midrule
1C  & 22 & $+0.033$ & 45 & $-0.018$ & $0.051$ & $0.001$  \\
2C  & 28 & $+0.045$ & 39 & $-0.010$ & $0.055$ & $<\!0.001$ \\
4C  & 30 & $+0.046$ & 37 & $-0.005$ & $0.051$ & $<\!0.001$ \\
8C  & 34 & $+0.051$ & 33 & $-0.006$ & $0.057$ & $<\!0.001$ \\
16C & 32 & $+0.045$ & 35 & $-0.007$ & $0.051$ & $<\!0.001$ \\
32C & 35 & $+0.024$ & 32 & $-0.005$ & $0.028$ & $<\!0.001$ \\
\bottomrule
\end{tabular}
\end{table}

\newpage
\section{Embedding space selection: TabClustPFN vs.\ ZEUS}
\label{app:vs_zeus}

We document the empirical basis for the unsupervised embedding model selection below, so the main results can be interpreted without ambiguity.

LUCoS replaces the original tabular feature space with a latent representation induced by an unsupervised Prior-Fitted Network. In our implementation, we use embeddings extracted from the PIN Encoder of TabClustPFN~\cite{zhao2026tabclustpfn}, a clustering-oriented PFN that produces 512-dimensional instance representations independently of the final cluster assignments. ZEUS~\cite{marszaek2025zeus} is an alternative unsupervised PFN that produces zero-shot embeddings optimized for class-boundary separation rather than cluster cohesion. Because the two models induce different latent geometries, we compare them empirically to assess which representation space better supports K-Medoids-based selection.

\paragraph{Experimental setup.}
ZEUS runs were terminated early for two reasons: preliminary results showed consistent underperformance relative to TabClustPFN, and inference frequently exceeded available GPU VRAM on larger datasets, preventing full-benchmark coverage.  Complete paired data exist for folds
1--5 on 58 of the 67 benchmark datasets (fold~0 yields only 4 ZEUS
datasets; folds 6--9 yield at most 1).  All comparisons are restricted
to this subset, giving 290 paired fold-level observations per budget
($58 \times 5$).  The KM\texttt{++} selector is held fixed throughout.
Differences are tested with two-sided Wilcoxon signed-rank tests; no
multiplicity correction is applied across budgets.
 
\paragraph{Results.}
Table~\ref{tab:embedding} reports mean AUC, W/L counts, and Wilcoxon
$p$-values per budget.  TabClustPFN is superior at every budget; every
comparison is significant ($p \leq 0.007$).  The gap peaks at
$2C$--$4C$ ($\Delta \approx +0.022$--$0.024$) and narrows
monotonically to $\Delta = +0.007$ at $32C$, consistent with
representation quality mattering most at the smallest labeling
budgets.
 
\begin{table}[ht]
  \centering
  \caption{Mean AUC of LUCoS under TabClustPFN and ZEUS embeddings,
    KM\texttt{++} selector, on 58 datasets with complete data
    (folds 1--5; 290 paired observations per budget).
    $\Delta = \text{TabClustPFN} - \text{ZEUS}$.
    W/L: fold-level pairs won/lost by TabClustPFN.
    $p$: two-sided Wilcoxon signed-rank test.
    All budgets significant; gap peaks at $2C$--$4C$ and narrows
    at $32C$ as increased data reduces dependence on representation
    quality.}
  \label{tab:embedding}
  \small
  \begin{tabular}{lcccrr}
    \toprule
    Budget & TabClustPFN & ZEUS & $\Delta$ & W/L & $p$ \\
    \midrule
    1C     & 0.656 & 0.646 & $+$0.010 &  167/96 & $\phantom{<}0.007$ \\
    2C     & 0.718 & 0.694 & $+$0.024 & 177/107 & $<$0.001 \\
    4C     & 0.770 & 0.748 & $+$0.022 &  197/91 & $<$0.001 \\
    8C     & 0.812 & 0.794 & $+$0.018 &  195/94 & $<$0.001 \\
    16C    & 0.851 & 0.835 & $+$0.016 &  187/99 & $<$0.001 \\
    32C    & 0.874 & 0.867 & $+$0.007 & 167/113 & $<$0.001 \\
    \midrule
    Global & 0.780 & 0.764 & $+$0.016 & 1090/600 & $<$0.001 \\
    \bottomrule
  \end{tabular}
\end{table}
 
\paragraph{Justification for TabClustPFN.}
The advantage is consistent and significant across all budgets.  At
$2C$--$4C$, choosing ZEUS costs approximately the same amount of AUC
as moving down one full budget level.  Mechanistically, this aligns
with the design objectives of the two models: TabClustPFN is trained
to induce clusterable geometry, making Euclidean distances used by
k-medoids more aligned with predictive similarity; ZEUS optimizes for
zero-shot class-boundary separation, a different objective that does
not guarantee the within-cluster homogeneity that medoid-based
coverage selection relies on.
 
The incomplete ZEUS coverage (58/67 datasets, 5/10 folds) is a
limitation of this comparison, arising from early termination of ZEUS
runs.  Given the consistency of the advantage across all six budgets
and 290 pairs per budget, we consider the evidence sufficient to
justify the choice of TabClustPFN as the default embedding for the
main experiments. This comparison is exploratory and is not used as evidence for the main LUCoS claim; it only supports the choice of TabClustPFN as the default representation.

\section{SSMA implementation} \label{sec:app_SSMA_implementation}

The SSMA algorithm used in this work is an adapted Python reimplementation of the original Java version proposed by its authors under the KEEL software kit \footnote{\url{https://github.com/SCI2SUGR/KEEL/blob/master/src/keel/Algorithms/Instance_Selection/SSMA/SSMA.java}}. The following sections describe the main components and modifications introduced in our implementation.

For this analysis, each training split is further divided into a candidate pool and a validation set, using 20\% of the training data for validation. SSMA searches over subsets of the candidate pool and optimizes their validation AUC when used as context for TabPFN-2.5, while also encouraging subset compactness. After optimization, the selected subset is evaluated on the held-out test split of the original fold.

At each iteration, two parents are selected via binary tournament selection. Variation operators (crossover and mutation) produce two offspring, which are subsequently evaluated. A local search procedure is conditionally applied to refine promising solutions. The population is updated using elitist replacement: the union of parents and offspring is sorted by fitness and truncated to retain the top-$P$ individuals. The process repeats until a maximum number of iterations is reached or a patience-based stopping criterion is triggered. The next subsections describe these steps in greater detail.

\subsection{Problem Formulation}
We address instance selection for tabular classification, aiming to reduce dataset size while preserving predictive performance. For each fold, $20\%$ of the training data is held out for validation, while the remaining $80\%$ is used as the candidate pool from which a reduced subset is selected. The algorithm optimizes this subset to maximize validation AUC under class coverage constraints. Candidate solutions are evaluated by using the selected subset as TabPFN context and combining validation AUC with a reduction objective into a scalar fitness.

\subsection{Chromosome Representation and Initialization}
Each solution is encoded as a binary vector over the candidate pool. It is initialized with 
$|S_0|=\min(0.25\cdot |D_{\mathrm{train}}|, 250)$ active instances, while enforcing at least one instance per class.

\subsection{Crossover}
Crossover is applied with probability $p_{\text{cross}}$ ($0.5$ in our implementation). When activated, uniform crossover is used: for each gene $j \in \{1,\dots,N\}$, offspring inherit the allele from one of the parents according to a Bernoulli mask. If crossover is not applied, offspring are direct copies of their parents. This operator does not explicitly enforce feasibility; constraint violations are handled during mutation.

\subsection{Mutation}
Mutation is implemented as a constrained bit-flip operator with an adaptive budget. The number of mutation events is sampled as
\[
n_{\text{mut}} = \sum_{j=1}^{N}\mathbf{1}[u_j < p_{\text{mut}}],
\]
and capped as
\[
n_{\text{mut}} \leftarrow \min\left(n_{\text{mut}}, 10, \left\lfloor 0.5 \cdot |S| \right\rfloor\right).
\]
Mutation events are divided into removals and additions with a $60/40$ split favoring removals. Additions activate inactive genes selected uniformly without replacement. After tentative modifications, a repair step ensures class coverage by activating at least one instance from any missing class.

Removals are applied sequentially. At each step, an active instance is considered \emph{protected} if it is the only representative of its class in $S$. Only non-protected instances are eligible for removal. This mechanism promotes reduction while preserving feasibility.

\subsection{Local Search}
We incorporate a stochastic first-improvement local search as a memetic refinement step. For a fixed number of attempts ($10$ in our implementation), the algorithm alternates between removal and addition moves. At each attempt, the removable set (excluding protected instances) and the inactive set are recomputed. With probability $0.7$, a removal is attempted if feasible; otherwise, an addition is performed. The selected gene is flipped, and the resulting solution is evaluated. The modification is accepted only if it strictly improves fitness; otherwise, it is reverted.

Local search is applied selectively, e.g., when offspring outperform the worst individual, with a small random probability (0.0625), or when $|S|$ is sufficiently small ($|S|<50$).

\subsection{Fitness Function}

The proposed formulation balances two competing objectives: maximizing validation AUC and minimizing the number of selected instances. To explore this trade-off, we optimize a scalar fitness function that combines both objectives through weighted aggregation. The reduction component is itself a non-linear function of the subset size, explicitly designed to maintain a strong incentive for further reduction even when the selected subset is already small, thereby avoiding premature convergence to overly large solutions.

Let $\text{AUC}_{\text{val}}(S)$ denote the validation AUC obtained by training TabPFN on $S$. The reduction objective combines linear and logarithmic components. The linear term is
\[
R_{\text{lin}} = 1 - \frac{|S|}{M}
\]

The logarithmic reduction term is designed to normalize subset sizes in log-space between a minimum feasible size ($C$; one instance per class) and a maximum allowed subset size ($M=500$). Specifically, we define
\[
R_{\log}^{\text{raw}} = 1 - \frac{\log_2 |S| - \log_2 C}{\log_2 M - \log_2 C}
\]
This formulation performs a min--max normalization of $\log_2 |S|$ between $\log_2 C$ and $\log_2 M$, ensuring that $R_{\log}^{\text{raw}} \in [0,1]$, with higher values assigned to smaller subsets. The resulting score is then rescaled to $[0.5,1]$ to reduce its relative dominance in the final fitness:
\[
R_{\log} = 0.5 + 0.5 \cdot R_{\log}^{\text{raw}}
\]
The combined reduction score is
\[
R = 0.5\,R_{\text{lin}} + 0.5\,R_{\log}
\]
The final fitness is
\[
F(S) = \alpha \cdot \text{AUC}_{\text{val}}(S) + (1-\alpha)\cdot R
\]
We set $\alpha=0.5$ for all experiments. Infeasible solutions violating class constraints are assigned zero fitness.

\subsection{Selection, Replacement, and Stopping}
Parent selection uses binary tournament selection. Replacement follows an elitist steady-state strategy: the current population and offspring are merged and truncated to size $P$ based on fitness. The algorithm tracks the best solution and maintains a set of non-dominated solutions that capture the trade-off between $|S|$ and predictive performance. 

The underlying objective of the genetic search is to progressively reduce the dataset size toward the theoretical minimum $|S| = C$, while preserving predictive performance. In cases where the algorithm stagnates at a larger subset size without further improvement, an additional diversification mechanism is triggered: individuals in the population are subjected to random pruning of selected instances to explicitly encourage further reduction and escape convergence to suboptimal large subsets.

Termination is determined either by a maximum iteration budget or by a patience criterion based on lack of improvement.

\end{document}